\def\year{2018}\relax
\documentclass[letterpaper]{article} 
\usepackage{aaai18}  
\usepackage{times}  
\usepackage{helvet}  
\usepackage{courier}  
\usepackage[hyphens]{url}  
\usepackage{graphicx}  
\usepackage{fancyhdr}
\usepackage{float} 
\usepackage{amsmath}
\usepackage{amssymb}
\begin{document}
%
\title{Towards a Unified Planner For Socially-Aware Navigation}
\author{Santosh Balajee Banisetty \dag \and David Feil-Seifer $*$\\
Computer Science \& Engineering\\
University of Nevada, Reno\\
1664 N.Virginia Street, Reno\\
Nevada 89557-0171\\
Email: santoshbanisetty@nevada.unr.edu \dag, 
dave@cse.unr.edu $*$
}
\maketitle
\begin{abstract}
This paper presents the framework for a novel Unified Socially-Aware Navigation (USAN) architecture and explains its need in Socially Assistive Robotics (SAR) applications. Our approach emphasizes interpersonal distance and how spatial communication can be used to build a unified planner for a human-robot collaborative environment. Socially-Aware Navigation (SAN) is vital for helping humans to feel comfortable and safe around robots; HRI studies have shown the importance of SAN transcends safety and comfort. SAN plays a crucial role in perceived intelligence, sociability and social capacity of the robot, thereby increasing the acceptance of the robots in public places. Human environments are very dynamic and pose serious social challenges to robots intended for interactions with people. For the robots to cope with the changing dynamics of a situation, there is a need to infer intent and detect changes in the interaction context. SAN has gained immense interest in the social robotics community; to the best of our knowledge, however, there is no planner that can adapt to different interaction contexts spontaneously after autonomously sensing the context. Most of the recent efforts involve social path planning for a single context. In this work, we propose a novel approach for a unified architecture to SAN that can plan and execute trajectories for an autonomously sensed interaction context that are human-friendly. Our approach augments the navigation stack of the Robot Operating System (ROS) utilizing machine learning and optimization tools. We modified the ROS navigation stack using a machine learning-based context classifier and a PaCcET based local planner for us to achieve the goals of USAN. We discuss our preliminary results and concrete plans on putting the pieces together in achieving USAN. 
\end{abstract}

\section{Introduction}
\label{sec:intro}
Socially assistive robotics (SAR)~\cite{feil2005defining} research demonstrates an increase in the efforts both in industry and academia to develop research and commercial applications to help people in real-world settings. Smart Luggage~\cite{Smartlauggage} showcased at 2018's consumer electronics show (CES) is a robotic suitcase that will follow the owner during travels. Some companies and start-ups develop robotic assistants for airports and shopping malls to assist people with directions and shopping experience. Robot domains, especially SAR, benefit from navigation as such movements extend the reachable service area of the robot. However, navigation, if not appropriately performed can cause an adverse social reaction~\cite{mutlu2008robots}. For socially-aware navigation (SAN), the spatial communication in which personal space and distance can be used to {\em communicate} with a human partner~\cite{feil-seifer2011people-aware}, which may increase in the acceptance of assistive robots in human environments. Spatial communication is one of the several subcategories of nonverbal communication. It can be defined as a study of the effects of spatial separation among individuals. It plays an essential role in everyday communication without people realizing it. For human-human interaction, humans can understand social norms via spatial communication without explicitly being told. For HRI to match this human-human interaction property, the spatial interface between a human and a robot should be utilized to achieve human-friendly navigation. Successful robot actions, including navigational actions, must be appropriate for a given social circumstance for the robot's long-term acceptance in human collaborative environments like hospitals, airports, and other public places.

Robots currently deployed in real-world settings have prompted negative reactions from people encountering them~\cite{businessinsider}. There can be many reasons why people are not liking these robots or not willing to interact with them, but particular reasons are unknown. One likely culprit is that the robots are not obeying social norms like getting in the way or treating people as mere obstacles and navigating around without considering personal space. In an ethnographic study \cite{mutlu2008robots}, one of the participants quoted the following statement: \\\\``\textit{Well, it almost ran me over... I wasn't scared... I was just mad... I've already been clipped by it. It does hurt.}''\\\\
Incidents such as these show the need for social path planning methods that involve the rules of proxemics are essential for successful human-robot interaction.    

Robots should not treat humans through their navigation behavior as dynamic objects \cite{trautman2010unfreezing}. Figures \ref{fig:results2} and \ref{fig:results} illustrate a comparison between a traditional planner (green trajectory), which optimizes for performance (time, distance, etc.) and a socially-aware planner (blue trajectory), which optimizes for social norms along with performance objectives. The traditional planner treats both people and objects the same way. Not able to behaviorally distinguish humans and objects can lead to HRI missteps; it is acceptable to get close to an object, but similar behavior around a person is not acceptable. Proxemics \cite{hall1966hidden} codifies this notion of personal space, yet there is ample evidence that human navigation preferences go beyond mere distance and includes motion \cite{mead2016autonomous}. Effective SAN will investigate methods to integrate the rules of interpersonal motion into robot navigation behavior. \citeauthor{kruse2013human} presented many methods in their survey that attempted to incorporate social norms in navigation planners and identifies the strengths and weaknesses of such approaches~\cite{kruse2013human}.

One limitation of the state-of-the-art SAN approaches is that current SAN approaches focus on only operating in a single scenario, but do not address differences between contexts. For example, a social planner that works in a ``meeting" context may not work in ``walking together" context. To address this limitation, we propose a combination of model and optimization-based approach that optimizes multiple navigation cardinal objectives for a sensed setting. For example, the objectives that are important for a detected context will be automatically selected and optimized to achieve socially-appropriate navigation behavior for that particular context. A unified SAN planner will only choose the cardinal objectives that matter most, as these objectives change from context to context. The primary contribution of this paper is the architecture, which assembles our previous work and ongoing work to achieve the objectives of USAN; we show results obtained so far and lay out concrete plans in realizing the proposed unified social motion planning architecture. Our SAN architecture, which is built on top of ROS navigation stack, will be modular and allows users to drop in own modules to configure a custom SAN method. 

The remainder of this paper is organized as follows. In the next section, we briefly discuss related work. Next, we describe the architecture of the proposed system and briefly go through the accomplished work so far. Later, we move on to future plans and discussion section.

\section{Challenges}\label{sec:challenges}

Robots operating alongside humans in a natural environment have to address particular challenges. Here, we will limit our discussion to challenges faced by robots due to their navigation behaviors and how social motion planners can address them. \citeauthor{kruse2013human} identified Comfort, Sociability, and Naturalness as challenges that SAN planners should tackle in a collaborative human environment \cite{kruse2013human}.

\textbf{C1 - Comfort.} \textit{Comfort is the absence of annoyance and stress or the easing of a person's feelings of distress in human-robot interactions.} 

\textbf{C2 - Sociability.} \textit{Sociability is communicating an ability or willingness to engage in social behavior and adherence to explicit high-level cultural conventions and social norms.} 

\textbf{C3 - Naturalness.} \textit{Naturalness is the low-level behavior similarity between humans and robots.} 

Adding to the above mentioned challenges, we would add safety, legibility, predictability, fluency, overall efficiency and acceptance as critical challenges faced by robots using traditional planners:  

\textbf{C4 - Safety.} \textit{Safety is the robots unlikeliness to cause harm or injury to human during integration.}

\textbf{C5 - Legibility.} \textit{Legibility is defined as the clarity in what a robot is doing and can a human identify it.}

\textbf{C6 - Predictability.} \textit{A bystander's ability to anticipate what the robot is currently doing and where it will go.}

\textbf{C7 - Fluency.} \textit{Fluency is the smoothness of the chosen trajectory.} \cite{hoffman2013evaluating}

\textbf{C8 - Overall Efficiency.} \textit{The combined task efficiency of both the robot and a human partner.}

\textbf{C9 - Acceptance.} \textit{Acceptance is the willingness of human to interact with the robot.}

One might think robots with traditional planners are safe as they avoid collisions, but these planners avoid humans within close proximity, which is not always safe. When it comes to the comfort of people interacting with robots, a robot with a traditional planner does not make people feel comfortable as it invades the personal space. Comfort and safety of humans around robots depend on legibility and predictability of motion taken by the robot to reach its goal. Predictability and legibility are fundamentally different and often contradictory properties of motion \cite{dragan2013legibility}. In human-human interaction, we use legibility and predictability of a human's trajectory to mutually avoid collision and invasion of personal space by understanding the intent of a person. Similarly, both legibility and predictability of the robot's trajectory help the human partners in understanding the robot's intended actions and vice-versa.

\section{Related Work}
Traditional navigation algorithms can generate a collision-free path and maneuver a robot on that path to get to a goal. However, these algorithms are not sophisticated enough to deal with social interactions that occur while navigating in highly dynamic human environments. There is a rapidly growing HRI community that is addressing SAN related challenges identified in the Challenges Section of this paper. The solutions to SAN associated problems range from simple cost functions to more advanced deep neural networks.

\citeauthor{ferrer2017robot} used the Social Force Model (SFM) \cite{helbing1995social} to mimic navigation behavior of humans. In \cite{ferrer2017robot}, the robot obeys the social forces while navigating to a goal. The method also extends the SFM to allow the robot to accompany a human while providing a method for learning the parameters of the model. \citeauthor{gomez2014fast} presented a special version of the fast-marching square planner to demonstrate social path planning \cite{gomez2014fast}. Furthermore, the authors proposed an extended mode to engage groups of people.  

\citeauthor{silva2017human} presented a Reinforcement Learning approach, where a robot learns a policy to share the effort required to avoid collision with a human \cite{silva2017human}. The results of the simulated experimental evaluation states that the robot mutually solves the collision avoidance problem with a human partner. \citeauthor{johnson2018socially} presented a SAN implementation on a smart wheelchair robot using topological map abstraction which lets the robot learn generalizable social norms \cite{johnson2018socially}. Furthermore, the authors compared their SAN planner with a baseline collision-free motion planner; the results show that a robot with SAN planner influenced the behavior of pedestrians around it.

Inverse Reinforcement Learning (IRL)-based planners can be scalable; however, they have limitations such as state space explosion and need for a significant amount of expert training data. \citeauthor{okal2016learning} presented a Bayesian Inverse Reinforcement Learning (BIRL) based approach to achieving socially normative robot navigation using expert demonstrations \cite{okal2016learning}. The authors extend BIRL to include a flexible graph-based representation to capture relevant task structure that relies on collections of sampled trajectories. \citeauthor{kretzschmar2016socially} proposed a method to learn policies from demonstrations; it learns the model parameters of cooperative human navigation behavior that match the observed behavior concerning user-defined features. They used Hamiltonian Markov chain Monte Carlo sampling to compute the feature expectations. To adequately explore the space of trajectories, the method relied on the Voronoi graph of the environment from start to target position of the robot~\cite{kretzschmar2016socially}.  \citeauthor{hamandi2018deepmotion} developed a novel approach using deep learning (LSTM) called DeepMoTIon, trained over well-known pedestrian surveillance data to predict human velocities \cite{hamandi2018deepmotion}. This work used a trained model to achieve human-aware navigation, where the robots imitate humans to navigate in crowded environments safely. 

\citeauthor{santana2018human} presented a human-aware navigation system for industrial mobile robots targeting cooperative intra-factory logistics scenario \cite{santana2018human}. The authors used cost functions to model assembly stations and operators in layered cost maps \cite{lu2014layered} to improve overall efficiency. \citeauthor{bordallo2015counterfactual} developed a multi-agent framework that utilizes counterfactual reasoning to infer and plan according to the movement intentions of goal-oriented agents~\cite{bordallo2015counterfactual}. \citeauthor{aroor2018online} formulated a Bayesian approach to develop an online global crowd model using a laser scanner. The model uses two new algorithms, CUSUM-A$^*$ (to track the spatiotemporal changes) and Risk-A$^*$ (to adjust for navigation cost due to interactions with humans), that rely on local observation to continuously update the crowd model~\cite{aroor2018online}.

\citeauthor{turnwald2018human} presented a game theoretic approach to SAN utilizing concepts from non-cooperative games and Nash equilibrium. The authors evaluated the game theory based SAN planner against established planners such as reciprocal velocity obstacles or social forces, a variation of the Turing test was administered which determines whether participants can differentiate between human motions and artificially generated motions~\cite{turnwald2018human}.

The work that exists deal only with a single context when addressing SAN, to the best of our knowledge, no method can handle multiple SAN contexts on the fly. \citeauthor{lu2014layered} work on layered costmaps is an approach that closely relates to the goals of USAN~\cite{lu2014layered}. However, it has limitations such as maintaining multiple costmaps can be memory intensive, computation of a master costmap from a subset of costmaps for a particular context can be computationally expensive. Also, this approach does not include a method to autonomously sense a context; hence, costmaps associated with a specific context cannot be selected automatically. On the other hand, IRL based approaches are promising in a single context and can be trained to handle multi-context SAN but will require a lot of human training data. The next section explains our approach towards a unified planner for socially-aware navigation which has a potential to overcome the said limitations.

\section{Technical Details}

\begin{figure}[t]
\centering
\includegraphics[height = 4.5cm]{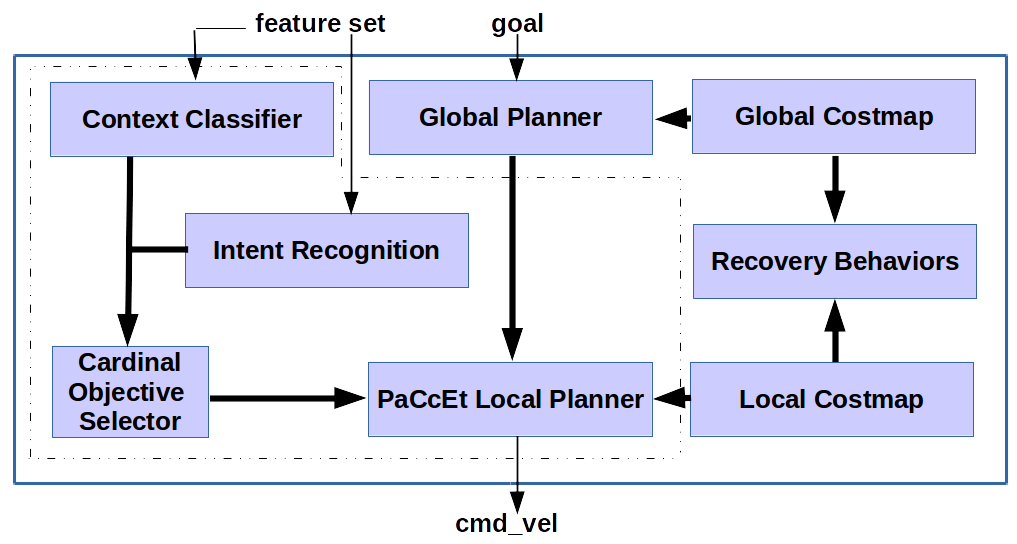}
\caption{An overview of the Unified Planner for Socially-Aware Navigation (UP-SAN). Modules with in the dotted lines are the modification to ROS navigation stack.}
\label{fig:usan}
\end{figure}

Prior work \cite{banisetty2016socially} demonstrated that actions could be distinguished from distance-based features using a model-based approach (GMM). In order to select objectives for a given context, we first need to autonomously sense the context of interaction (passing, meeting, etc.). We modified the ROS navigation stack and demonstrated in a simulation that a PaCcET based local planner~\cite{scott} can achieve better performance concerning human experience when compared to a traditional local planner. With our proposed approach, we will address the limitation identified in \textit{Introduction} Section by realizing a navigation stack as shown in Figure~\ref{fig:usan}. The global planner, computation of costmaps, and recovery behaviors of the original navigation stack were untouched as they are well implemented and does the job. For example, the recovery methods implemented in original ROS navigation stack also holds well in our modified stack. The blocks enclosed in dotted lines represent our novel concepts which will be added to the ROS navigation stack, detailed in the following sub-sections. First, we will detail a context classifier that autonomously detects the navigation context. Next, we discuss the role of an intent recognition system. Next, we detail a procedure using which the robots selects the cardinal objectives for the detected context and then explain how PaCcET (Pareto Concavity Elimination Transformation) framework~\cite{yliniemi2014paccet} is used at a low-level planning task. Our idea is to develop a customizable architecture that can be modified, for example, one can replace context classifier model with a custom model of the contexts that the researchers are interested in.

\subsection{Context Classifier}
A unified socially-aware navigation method must dynamically sense interpersonal and environmental features to identify a context. For this purpose, we previously employed a model-based method to determine the context based on a feature set (environmental features like walls, doorways, etc. and distance based features like interpersonal distance, distance from a wall, etc.). We can use any machine learning classifier depending on the complexity of the contexts. In one of our prior works \cite{banisetty2016socially}, we used Gaussian Mixture Models (GMM) to implement the context classification functionality trained on a set of distance-based features. Our GMM model was able to distinguish between different scenarios (Passing, Meeting, Walking together towards a goal, and Walking together away from a goal) with an accuracy of 94.74\%. However, we are currently investigating the applications of newer classification techniques that can scale with data. While the GMM-based model demonstrated good classification accuracy for different scenarios in a hallway context, it will not scale to other contexts like art gallery/museum interactions, joining a group, etc. Unlike GMMs, one advantage of neural nets is that we do not have to hand pick the features and hence is an ideal choice for context classification and scene understanding. We are investigating neural net-based perception methods that can classify complex scenarios like art gallery interaction, joining a group, etc.  

\subsection{Intent Recognition}
The environments that SAR will be deployed are complex and involves other decision-making agents such as other robots or humans. In environments like public places, a reactive social planner will only yield a sub-optimal human-robot interaction experience. These complex environments call for an intent recognition module integrated into the planning pipeline as shown in Figure~\ref{fig:usan}. Intent recognition in SAN is not new, researchers in recent times have explored it~\cite{kelley2008understanding}. However, intent recognition in a unified planning architecture is not only novel but also a key component in achieving the objectives of USAN.

SAN problem to some extent boils down to a multi-agent optimization problem. When all the agents in a multi-agent system are robots, it is easy to solve as we have similar sensors and standard communication protocols. However, when robots are in a human environment, there is no such standard communication protocol between people and robots. On the other hand, when people interact, we utilize spatial communication to infer the intent of others. Similarly, there is a need for an intent recognition system that can understand human navigation intentions in a human-robot interaction scenario. Our group is developing an intent recognition module that utilizes OpenPose \cite{cao2017realtime}. The intent recognition system will answer questions, such as:
\begin{itemize}
    \item Do particular people belong to a group?
    \item Does a person belongs to a waiting queue or he/she is just standing talking to another person?
    \item Is someone interested in interaction?
    \item Are the group dynamics changing? 
\end{itemize}

\subsection{Cardinal Objectives Selector}
\label{sssec:selector}
In our prior work \cite{banisetty2016socially}, we used GMM to select the future trajectory points using the probability score of the classification label provided by the GMM. The probability score that $w$ is part of the model and is given by the following probability density equation:

\begin{equation}
\label{pdf}
p(w, k|\phi) = \frac{1}{(2\pi)^{n/2}|\sum_{\phi(k)}|^{1/2}} e^{- \delta_M(w, k|\phi)}
\end{equation}


Other methods also provide a probability score, such as neural nets. Depending on the complexity of the environment, the complexity of scenarios, and the available computational resources; we can choose one over the other (different classification methods). For the Cardinal Objective Selector, the context classifier provides a probability distribution over any potential navigation scenarios. We can use this probability score in selecting the cardinal objectives $(Obj_1, .. , Obj_n)$ or emphasize on how much each cardinal objective $(w_1*Obj_1, w_2*Obj_2, .. , w_n*Obj_n)$ is important based on the context. Where, $w_i$ is the probability score obtained using Equation~\ref{pdf}. Using PaCcET local planner, we are able to optimize for a subset of objectives that matter most for a sensed context. This will help filter out unnecessary factors in an interaction as well as speed computation to meet real-time constraints.  

\subsection{PaCcET Local Planner}
For every future trajectory point, the traditional local planner in ROS navigation stack minimizes the following cost function, shown in Equation \ref{eq:Cost function}. This cost function does not make use of any social features associated with the future trajectory points. The traditional method is a linear combination of weighted objectives fitness scores. In reaching a goal, the traditional method can lead to an optimal set of policies concerning getting to the goal as quickly as possible, disregarding any social considerations. However, a SAN planner may yield sub-optimal policies (using traditional approaches) in order to prioritize social considerations. A solution to this is to use a multi-objective tool like PaCcET to evaluate policies on multiple objectives \cite{yliniemi2014paccet,yliniemi2015complete} properly. For a hallway context, multiple objectives can be to navigate on the right side, maintain an appropriate distance from other people in the hallway, etc.

\begin{equation}
\begin{split}
\label{eq:Cost function}
\text{cost}({v}_{x},{v}_{y},{v}_{\theta}) &= \alpha(\Delta_{path})+\beta(\Delta_{goal})+\gamma(\Delta_{heading})\\
& +\delta(\Delta_{occ})
\end{split}
\end{equation}


Unlike the traditional approaches, our modified PaCcET local planner uses a multi-objective optimization approach to minimize the cost function as shown in Equation \ref{eq:PaCcET fitness example}. Here, $Obj_i$ is a social feature and will be selected (based on the context) as discussed previously. Unlike other modules in USAN architecture, PaCcET based local planner takes low-level decision related to proxemics and the high-level decision like type of interaction, selection of objectives happen in other modules as discussed in earlier sections. 

\begin{align}
\label{eq:PaCcET fitness example}
\text{$P_f$}= T_f(cost({v}_{x},{v}_{y},{v}_{\theta}),Obj_1,Obj_2,....,Obj_n)
\end{align}

For low-level planning task, PaCcET is used over the other multi-objective tools because of its computation speed \cite{yliniemi2014paccet}. PaCcET can be used to evaluate the possible trajectories developed in the local planner. At each time step, the sensor data are analyzed, and the desired features are evaluated for each of the potential future trajectory point. PaCcET then uses the fitness values for each feature of every future trajectory point to develop the solution space and obtain the optimal future trajectory. At every time step, a future trajectory point is generated, PaCcET outputs a Pareto solution for each such generated point. By using PaCcET on every trajectory point, the local planner can be optimized for social norms in real time. In contrast, a conventional planner uses a simple combination of scores for distance, goal, etc. 

\begin{figure}[t]
\centering
\includegraphics[height = 4.5cm]{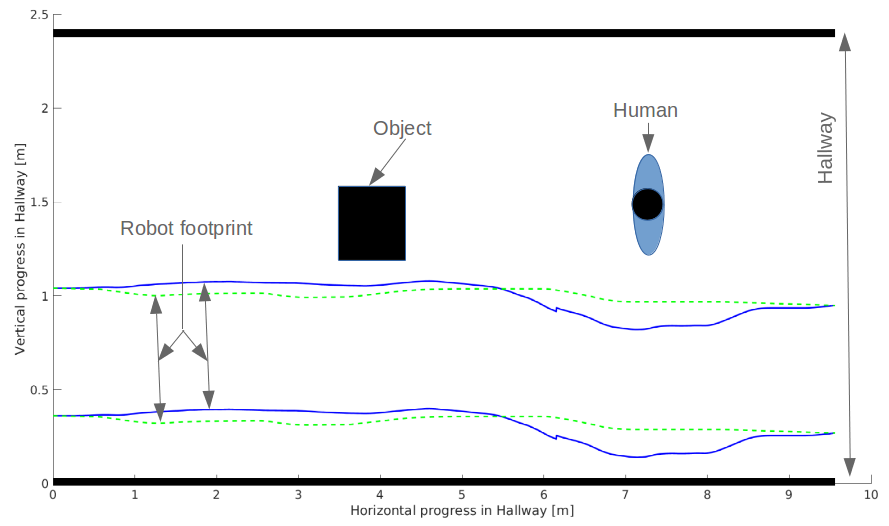}
\caption{Showing PaCcET local planner in comparison with traditional local planner that does not account for social norms. Traditional planner generated a trajectory that is close to both human and the object (black box), treating them alike. Our approach, PaCcET based SAN planner, generated a trajectory that diverged around the human, thereby respecting the personal space of the human.}
\label{fig:results2}
\end{figure}

\begin{figure}[t]
\centering
\includegraphics[height = 4.5cm]{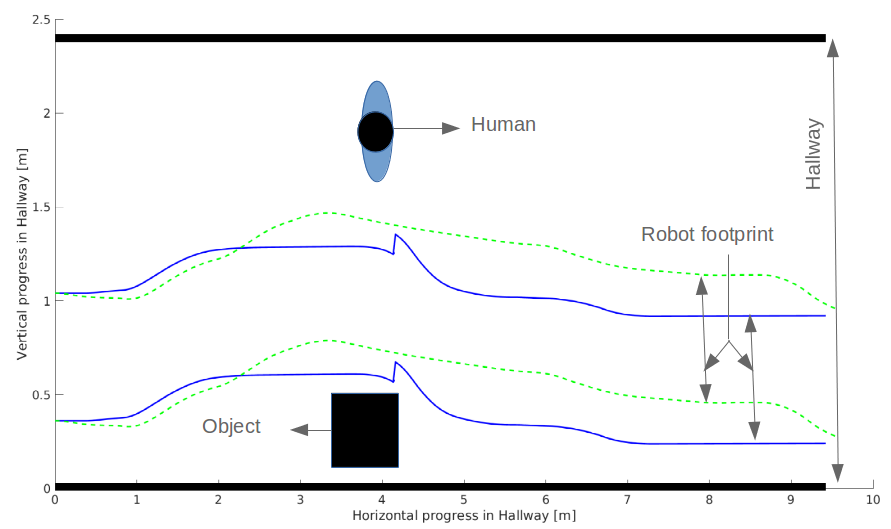}
\caption{Showing PaCcET local planner in comparison with traditional local planner in a tight navigation situation. Our approach (blue trajectory) was able to distinguish between a person and an object by its navigation behavior, where as the traditional planner (green trajectory) navigated the robot more centrally.}
\label{fig:results}
\end{figure}

In \cite{scott}, we demonstrated that PaCcET based local planner could perform better than a traditional planner regarding goals associated with SAN. Figures \ref{fig:results2}, \ref{fig:results} show two such results not covered in \cite{scott}. Figure \ref{fig:results2} shows that PaCcET local planner (blue trajectory) when around an object (black box) optimizes for shortest distance and does not deviate from the global path. However, when around a human, it also considers personal space and deviates from the global path so that the robot does not get into the human's personal space. Conversely, traditional planner (green trajectory) treats both the human and the object as an obstacle merely to avoid collisions, but without additional deference to the human; this behavior of treating both objects and people as mere obstacles is not socially appropriate. 

Figure \ref{fig:results} shows a scenario where there is a tight space between a person and an object; in this case, the person is not interacting with the object. If the person is interacting with the object, getting in between without asking for an excuse is inappropriate. In this situation (see Figure \ref{fig:results}) the PaCcET local planner (blue trajectory) guides the robot to reach its goal by getting close to the object thereby leaving more space around the person. The SAN planner made the robot move a little towards the person (seen as a jerk in motion) as to avoid running into the object on its right (using its holonomic movements). This behavior is due to space constraint in the hallway as the robot is trying to reach its goal while avoiding invasion of the space around the person. However, the traditional planner (green trajectory) makes the robot navigate more centrally, treating both person and obstacle alike, which is socially inappropriate.  


\section{Extending to Multiple Contexts}
\begin{figure}[t]
\centering
\includegraphics[height = 8cm]{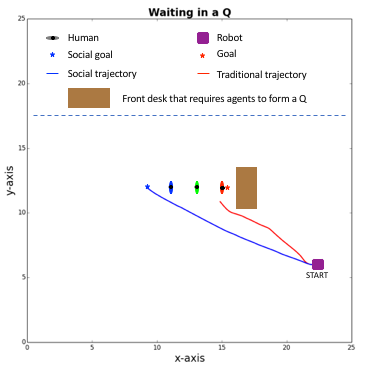}
\caption{Showing PaCcET local planner (blue trajectory) in comparison with traditional local planner (red trajectory) in a ``Waiting in a queue scenario". Our approach steered the robot to join the line, whereas the traditional planner steered the robot to the front of the desk.}
\label{fig:newresults2}
\end{figure}

While Figures \ref{fig:results2} and \ref{fig:results} show how PaCcET local planner behave more appropriately from a social standpoint than a traditional planner, both scenarios involve a single person in a hallway with only two objectives to optimize. In ongoing work, we extended our local planning concept to different scenarios/contexts like a robot waiting in a queue, robot joining a group and robot presenting to the audience in an art gallery. The result shown in Figure \ref{fig:newresults2} is an example to demonstrate PaCcET local planner in a much complex context involving more humans and more objectives. 

In ``waiting in queue" context as shown in Figure~\ref{fig:newresults2}, both PaCcET and traditional planners were given the same start (shown as START) and goal (denoted by a red star, in front of the person in a red shirt) positions. Here, we chose a front desk interaction in an office, but this can be generalized to similar situations which involve agents (both humans and robots) to form a line to get to a resource. For example, interaction at a public coffee machine or a vending machine. The trajectory executed by the traditional planner tried to reach the goal without any social considerations, thereby cutting the line and positioning the robot in a socially inappropriate location. On the other hand, the PaCcET local planner instead of getting to the goal directly, it steered the robot towards a social goal (the end of the line in this context). For the demonstration of the low-level capabilities of our PaCcET local planner, the social goal was hand-picked. An automated social goal detection can be achieved as a high-level decision by fitting a straight line (in this case) with all the detected people in the scene and calculating a point at the end of the line considering proxemics around the last person. Social goal location changes based on context, for example, in a situation, where the robot is required to join a group of people, it should position itself in such a way that it can interact with everyone in the group. In this case, social goal computation can be achieved by finding a location in the circle that can be formed by the group (referred as O-formation in literature). 

\section{Future Work and Discussion}
Currently, we are working on integrating PaCcET local planner with a context classifier and an intent recognition system. Once we have the proposed system implemented as shown in Figure \ref{fig:usan}, our planner can be used on any omni-drive or differential drive robots, compatible with the ROS navigation stack. To validate our planner on the claimed platforms, we will implement and test our planner on PR2 (omni-drive) and Pioneer (differential drive) robots. We are also working on implementing layered costmaps approach~\cite{lu2014layered} on a common scenarios to compare with our proposed approach. 

The new planner on these two platforms can be evaluated in the following two ways (in comparison to the traditional planner and a layered cost map approach or an openly available socially-aware planner):
\subsection{Performance of the planner}
The performance of the planner will be evaluated using metrics like time, distance, distance maintained, etc. as discussed in \cite{sebastian2017socially} and other similar metrics like number of proxemic intrusions, etc.
\subsection{Social effects of the planner}
Social aspects of the planner will be evaluated as detailed in the sections below:
\subsubsection{In-person experiments}
We will conduct a 2 x 2 between-participant study with an interacting partner (human or robot) and navigation type (traditional navigation behavior or SAN behavior). We plan to recruit 40 participants (10 per cell). Pre- and post-questionnaires will be answered by the participants that will include validated sub-scales related to the social priorities of SAN (Comfort, Naturalness, Appropriateness, Sociability, etc.).

\subsubsection{Observer experiments}
We will utilize the navigation behavior from the prior study, represented as Heider \& Simmel-style videos \cite{heider1944experimental} that preserve the spatial relationship between the human(s) and robot while obscuring whether the agents are humans or robots. Observers will be asked to rate on a 7-point Likert scale the appropriateness on the sub-scales described above.

\subsection{Metrics}
The list identified in the Challenges Section needs metrics for us and the community to test SAN planners. Keeping in mind, the growing SAN research community, we are working towards identifying both qualitative and quantitative metrics to measure human-robot interaction quality during navigation tasks. When measuring social aspects of such navigation behaviors or any robot behaviors in general, we identified a few aspects that are missing from standardized surveys~\cite{bartneck2009measurement,syrdal2009negative}. We are currently working with psychometricians on perceived social intelligence (PSI) scales that can be used by the HRI community~\cite{barchard2019perceived}.

\section{Conclusion} 
\label{sec:conclusion}
In this paper, we clearly explained the need for a unified SAN architecture; we showed how a context classifier, an intent recognition system, a cardinal objective selector and a modified planner could achieve the goals of a unified SAN planner. Prior work on GMM based action discrimination was able to classify ongoing interactions as meeting, passing, walking together towards a goal and away from a goal. We discussed the limitations of our GMM based approach, discussed how a Neural Net based perception module can classify complex contexts. Another work, PaCcET based local planner was able to demonstrate that taking into account spatial features in multi-objective optimization problem can yield socially appropriate trajectories for a simple, single person interaction in a hallway. Ongoing efforts extend PaCcET local planner to handle multiple features which can be helpful in feature-rich interactions like group interactions or complex human environments. Our preliminary results in the individual subsystems show that our architecture has a potential to address the limitation of other SAN planners.

The literature we found dealt with a single context, it is worth investigating the proposed approach which can steer the community towards a Unified Socially-Aware Navigation (USAN).

\section*{Acknowledgments}
The authors would like to acknowledge the financial support of this work by the National Science Foundation (NSF, \#IIS-1719027), Nevada NASA EPSCoR (\#NNX15AK48A), and the Office of Naval Research (ONR, \#N00014-16-1-2312, \#N00014-14-1-0776). 

\bibliographystyle{aaai}
\bibliography{references}

\end{document}